# A Comparative Analysis on the Applicability of Entropy in remote sensing

S.K. Katiyar  Arun P.V[1]

Dept. Of Civil
MANIT-Bhopal, India
Ph: +914828149999

**Abstract**

Entropy is the measure of uncertainty in any data and is adopted for maximisation of mutual information in many remote sensing operations. The availability of wide entropy variations motivated us for an investigation over the suitability preference of these versions to specific operations. Methodologies were implemented in Matlab and were enhanced with entropy variations. Evaluation of various implementations was based on different statistical parameters with reference to the study area The popular available versions like Tsalli's, Shanon's, and Renyi's entropies were analysed in context of various remote sensing operations namely thresholding, clustering and registration.

## I. Introduction

Entropy as a measure of uncertainty associated with information was introduced by Claude E. Shannon being inspired from the common entropy concept in physics (Long et al, 2000). The principle of entropy is to use uncertainty as a measure to describe the information contained in a source (Robert et al, 1999). In information theory, the concept of entropy is used to quantify the amount of information necessary to describe the macro state of a system (Frieden et al, 1972). Different versions of entropies have been applied for the effective automation of various remote sensing analyses (Burch et al, 1983; Zhuang et al, 1987; Medha et al, 2009; Arun et al, 2012) and hence it is needed to investigate the suitability preference of specific entropy versions to various operations. The entropy is related to the concept of Kolmogorov complexity, which reflects the information content of a sequence of symbols independent of any particular probability model (Zhuang et al, 1987). More specifically, the Kolmogorov complexity of an object is a measure of the computational resources needed to specify the object (Shenet al, 1999). If one has no preference among samples resulting from an experiment, the best decision is not to introduce any biased knowledge into the decision process. Instead, all samples must be treated equally important (Shen et al, 1987). In this case, the probability distribution that describes the experiment is either uniformly distributed in continuous probability space or equally likely in discrete probability space, both of which yield the ME (Shanon et al, 1987). In this context, the term usually refers to the Shannon entropy, which quantifies the expected value of the information contained in a message, usually in units such as bits (Li et al, 1993).

---

[1] Email:arunpv2601@gmail.com

Shannon entropy is a measure of the average information content when one does not know the value of the random variable (Cover et al, 1991). The maximum information is achieved when no a priori knowledge is available, in which case, it results in maximum uncertainty. In Shannon information theory, the entropy is a measure of the uncertainty over the true content of a message, but the task is complicated by the fact that successive bits in a string are not random, and therefore not mutually independent, in a real message (Shanon et al, 1948). The two-dimensional Tsallis entropy (Tsallis et al, 1948) is obtained from the two dimensional histogram which is determined by using the gray value of the pixels and is applied as a generalized entropy formalism (Mohamed et al, 2011). Previously, entropy has been a metric difficult to evaluate without imposing unrealistic assumptions about the data distributions (Renyi et al, 1960). Renyi's entropy lends itself nicely to non-parametric estimation, overcoming the difficulty in evaluating traditional entropy metrics (Sahoo et al, 1997).

The various entropy variations can be applied for the enhancement of various remote sensing operations namely thresholding, registration and clustering and hence is needed to investigate the suitability preference of various versions. Thresholding is an important technique in image processing tasks and many methods for the same are found over the literature (Li et al, 1993). Information theoretic approach based on the concept of entropy considers image histogram as a probability distribution, and then selects an optimal threshold value that yields the ME (Chang et al,2006). ME based thresholding was first proposed by Pun et al (2006) and was later improved by Kapur et.al (1985) that was further generalised to Renyi's entropy (Renyi et al,1960). Pun and Kapur et al. methods are considered as first-order entropy thresholding method where as Abutaleb's (1989) cooccurrence based method and Pal's Joint Entropy - Local Entropy based method (2003) are thought of as second-order methods. The crucial difference between entropy thresholding and relative entropy thresholding is that the former maximises Shannon's entropy, whereas the latter minimises relative entropy.

Image registration is the process of calculating the spatial geometric transforms that aligns a set of images to a common observational framework (Barbara et al, 2003). The feature matching step in automation of image registration algorithms may be enhanced by adopting the maximization of mutual information. Mutual information enables to extract an optimal match with a much better accuracy than cross-correlation, and can be applied successfully to the registration of remotely sensed imagery (Medha et al, 2009). The concept of mutual information represents a measure of relative entropy between two sets, which can also be described as a measure of information redundancy (Antoine et al, 1998). A lot of mutual information based methods are available in literature (Van Den et al, 1998) and an optimal one suggested by Arun et.al is selected from among them to have a comparative analysis for the entropy techniques (Katiyar et al., 2012).

Clustering is one of the fundamental problems of pattern recognition that organizes the data patterns into natural groups, or clusters, in an unsupervised manner (Robert et al, 2003). Entropy can be used as similarity measure which enables to utilize all the information contained in the distribution of the data, and not only the mere second order statistics as in many traditional algorithms (Robert et al, 1999). Groupings can be evaluated by quantifying the entropy as cluster evaluation function (CEF) which was first introduced by Gokcay et.al (Gokcay et al, 2002). The use of entropy to the clustering function enables to find the clusters of any shape, without knowing the true number of clusters in advance. We have adopted an entropy based clustering algorithm developed by Jenssen et.al for the comparative analysis (Jenssen et al, 2002).

In this paper we have analysed the comparative suitability of the various available entropy versions in context of a few remote sensing analyses namely image thresholding, image registration and, image segmentation. The major entropies like Tsallis, Renyis and Shanon entropies were compared in the context of above operations to analyze the suitability of specific versions to specific operations.

## II. Methodology: Comparative analysis of different algorithms

The Satellite images namely LISS III and LISS IV sensor images of Bhopal area were used as test images for comparing the performance of various entropy enhanced implementations. The algorithms were modified to enhance them with entropies and to facilitate the variation of entropy. The various algorithms were implemented in MATLAB and the accuracies were computed using ERDAS imagine and the results were compared. The suitability of entropy versions for registration process was analysed by adopting the method developed by Arun et.al (Arun et al,2012) and accuracy were estimated using the following criteria: normalized cross-correlation coefficient (NCCC) (Medha et al,2009) and root mean square error (RMSE) (Antoine et al,1998). Prior to registration, in order to make same resolution, high resolution images were degraded to match with coarse spatial resolution data.

The comparative analysis of the different entropies in context of image segmentation were analysed based on method developed by Jenssen et.al (2002) and various statistical parameters namely Kappa statistics and over all Accuracy were computed for accuracy analysis. The accuracy of image thresholding implementations were evaluated by adopting the cross correlation coefficient (Van Den et al,1998) and average score (Mehmet et al,2004) estimation between threshold image and original one. The ground truthing for various accuracy estimations were done with reference to the Google earth and Differential Global Positioning System (DGPS) survey data over the Bhopal city and surrounding areas. The methodology adopted for this research work is as given in (Figure 1).

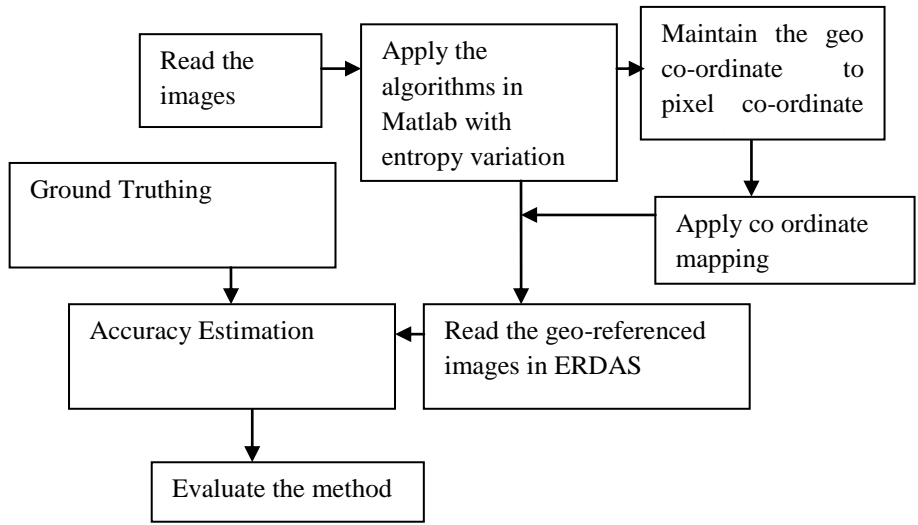

Figure 1. Methodology Adopted

## III. Results and discussion

The comparative analysis of various entropies over image registration techniques have been verified on various satellite images as LISS III, LISS IV and the results of observations are as summarised in the (Table 1). The Normalized Cross Correlation Coefficient measures the similarity between the images. The NCCC value ranges from (0-1) and an NCCC value of unity indicates perfectly registered images. The RMSE value indicates the error in registration and a least RMSE value is preferred for a perfect registration. The execution time was also analyzed using MATLAB counter function and generally categorized as high (>60 sec), medium (30-60 sec), low (<30 sec). The results presented confirms that the Renyi's entropy is more preferable followed by Tsalli's when compared to their counterparts.

Table 1. Accuracy comparison of registration

| S.No | TEST DATA | | TECHNIQUE | NCCC | RMSE | EXECUTION TIME |
|---|---|---|---|---|---|---|
| | Master Image | Slave Image | | | | |
| 1 | LISS3 | LISS4 | Tsallis | 0.65 | 4.12 | High |
| 2 | LISS3 | LISS4 | Renyi's | 0.67 | 3.81 | Low |
| 3 | LISS3 | LISS4 | Shanon | 0.53 | 4.82 | Higher |

The investigations on the suitability of entropy versions for image segmentation revealed that Tsallis entropy is preferable followed by Renyi's and Shanon. The efficiency of the Entropy variations were evaluated with reference to the Jessen et.al clustering and accuracy in terms of various statistical measures are as summarised in (Table 2). The increase in value of kappa statistics (maximum = 1) and overall accuracy (max=100%) indicates betterness of the methodology.

Table 2. Accuracy Comparison

| S.No | Sensor | Methodology (Entropy adopted) | Kappa statistics | Overall Accuracy (%) |
|---|---|---|---|---|
| 1 | LISS 3 | Tsallis | 0.96 | 96.83 |
| 2 | LISS 3 | Renyi's | 0.93 | 94.58 |
| 3 | LISS 3 | Shanon | 0.92 | 93.13 |
| 4 | LISS 4 | Tsallis | 0.94 | 94.80 |
| 5 | LISS 4 | Renyi's | 0.91 | 93.00 |
| 6 | LISS 4 | Shanon | 0.90 | 91.00 |

The comparative analysis of entropy techniques in the context of thresholding revealed that Tsallis entropy is comparatively more suitable when compared to the Renyi's and Shanon. The results are as summarised in (Table 3) and the average score, correlation coefficients were used for accuracy estimation. Greater value of correlation and average score corresponds to better thresholding technique.

Table 3. Accuracy Comparison

| S.No | Sensor | Methodology (Entropy adopted) | Average Score | Correlation Coeff. |
|---|---|---|---|---|
| 1 | LISS 3 | Tsallis | 1.11 | 0.78 |
| 2 | | Renyi's | 0.60 | 0.64 |
| 3 | | Shanon | 0.02 | 0.51 |
| 4 | LISS 4 | Tsallis | 1.01 | 0.81 |
| 5 | | Renyi's | 0.94 | 0.68 |
| 6 | | Shanon | 0.03 | 0.74 |

## IV. Conclusion

The investigations over the suitability of different entropy techniques in the context of various remote sensing operations revealed that Renyi's entropy is suitable for image registration purpose followed by Tsalli's and Shanon where as Tsalli's entropy is found preferable for thresholding and clustering. Renyi's is the simplest entropy method and is computationally simple as it avoids parametric estimation. The various experiments revealed that Shanon is the computationally complex entropy variation and hence is less preferable in the above discussed operations. Since Renyi's and Tsalli's are a limiting case of Shanon for a better accurate estimation Shanon may be adopted.

## V. References

Abutaleb A.S.,1989, "Automatic thresholding of gray-level pictures using two-dimensional entropy," Comput. Vision Graphics Image Process, 47: 22-32.